\documentclass[conference]{IEEEtran}
\IEEEoverridecommandlockouts
\usepackage{cite}
\usepackage{amsmath,amssymb,amsfonts}
\usepackage{algorithmic}
\usepackage{graphicx}
\usepackage{textcomp}
\usepackage{xcolor}
\usepackage{algorithm}
\usepackage{algorithmic}
\usepackage{subcaption} 

\def\BibTeX{{\rm B\kern-.05em{\sc i\kern-.025em b}\kern-.08em
    T\kern-.1667em\lower.7ex\hbox{E}\kern-.125emX}}
\begin{document}


\title{Deep Unfolding for MIMO Signal Detection}



\author{
    \IEEEauthorblockN{
        Hangli Ge\IEEEauthorrefmark{1}, Noboru Koshizuka\IEEEauthorrefmark{1},\\
    }
    \IEEEauthorblockA{
       \IEEEauthorrefmark{1}\textit{Interfaculty Initiative in Information Studies, The University of Tokyo}\\
        Email: \{hangli.ge, noboru\}@koshizuka-lab.org
    }
}

\maketitle

\begin{abstract}
In this paper, we propose a deep unfolding neural network-based MIMO detector that incorporates complex-valued computations using Wirtinger calculus. The method, referred as Dynamic Partially Shrinkage Thresholding (DPST), enables efficient, interpretable, and low-complexity MIMO signal detection. Unlike prior approaches that rely on real-valued approximations, our method operates natively in the complex domain, aligning with the fundamental nature of signal processing tasks. The proposed algorithm requires only a small number of trainable parameters, allowing for simplified training. Numerical results demonstrate that the proposed method achieves superior detection performance with fewer iterations and lower computational complexity, making it a practical solution for next-generation massive MIMO systems.
\end{abstract}


\section{Introduction}
Multiple-input multiple-output (MIMO) is a crucial component of modern in wireless communications\cite{ yang2015fifty,albreem2019massive}. A practical detector for massive MIMO systems must not only achieve detection performance but low computational complexity. Most traditional MIMO detection algorithms have difficulties with balancing on both metrics.
In recent years, with advancements in artificial intelligence, neural networks (NNs) have been proposed in improving signal processing tasks in MIMO systems. Deep learning has shown promise in improving detection accuracy through approximation. However, these models have notable drawbacks, including their ``black box" nature limits interpretability or explainability. Additionally, they require large datasets and extensive training time to perform effectively. When deploying at the physical layer, such models face practical challenges due to limited memory and computational power, making current neural network-based solutions difficult to apply in real-world scenarios\cite{balatsoukas2019deep,jagannath2021redefining}.


To address these challenges, deep unfolding methods \cite{gregor2010learning} have been proposed.
These approaches aim to find accurate solutions using deep learning principles, while requiring only a small number of iterations. Deep unfolding techniques involve minimal parameter tuning (e.g., step-size selection), offering a more interpretable and computationally feasible alternative for MIMO detection. This concept of data-driven tuning in numerical optimization algorithms originates from the work of Gregor and LeCun \cite{gregor2010learning}. Building on this idea, we propose a deep unfolding neural network-aided MIMO signal detector that operates in the complex domain. Our approach introduces a mathematical formulation based on complex-valued computations, utilizing Wirtinger calculus to derive gradients.Our proposal based on a solid mathematical foundation that is intuitive, interpretable, and has low complexity and memory requirements. It enables a relatively simple and principled methodology for solving such optimization problems.

\section{Deep Unfolding Neural Network-aided Methods}

\subsection{Problem Definition}

Consider a MIMO (Multiple-Input Multiple-Output) communication system with \(N_t\) transmit antennas and \(N_r\) receive antennas. The received signal vector \(\mathbf{y} \in \mathbb{C}^{N_r \times 1}\) can be expressed as:
\begin{equation}
    \mathbf{y} = \mathbf{H} \mathbf{x} + \mathbf{n},
\end{equation}
\begin{itemize}
    \item \(\mathbf{y}\ \in N_r \times 1\) is the complex-valued received signal vector,
    \item \(\mathbf{H} \in \mathbb{C}^{N_r \times N_t}\) is the channel matrix between \(N_t\) transmit and \(N_r\) receive antennas.
    \item \(\mathbf{x} \in \mathbb{C}^{N_t \times 1}\) is the \(N_t \times 1\) the complex-valued transmitted signal vector,
    \item \(\mathbf{n} \in \mathbb{C}^{N_r \times 1}\) is the noise vector, typically modeled as complex Gaussian vector distributed according to \((0, \sigma^2 \mathbf{I}_{N_r})\).
\end{itemize}
The entries \(h_{ij}\) of \(\mathbf{H}\) are often modeled as independent and identically distributed (i.i.d.) complex Gaussian random variables (Rayleigh fading) with zero mean and variance \(\sigma^2\). It can be expressed as:\begin{equation}
    h_{ij} \sim \mathcal{CN}(0, \sigma^2\mathit{I}),
\end{equation}

where \(\mathcal{CN}(0, \sigma^2\mathit{I})\) denotes a circularly symmetric complex Gaussian distribution with mean 0 and variance \(\sigma^2\mathit{I}\). The goal of MIMO sigal detector is as defined as Equation~\ref{opt}, to solve the following optimization problem:
\begin{equation}
\begin{aligned}
    & \underset{\mathbf{\widetilde{x} \in \mathbb{C}^{N_t \times 1}}}{\text{minimize }}\mathbf{\|H\widetilde{x}-y}\|_2^{2}
\end{aligned}
\label{opt}
\end{equation}

where $\|\|_2^{2}$ represents the $\displaystyle \ell ^{2}$ (Euclidean) norm. 



\section{Proposed Method}\label{proposal}


The proposed method is called Dynamic Partially Shrinkage Thresholding (DPST), which is an iterative algorithm with a fixed number of iterations of $\mathit T$, unfolds its structure, and introduces a number of trainable parameters (denoted as params). $\mathit T$-step iterative inference algorithm can be unfolded into an $\mathit T$-layered neural network (NN) with trainable parameters based on the model.
The objective function was extended in equation \ref{eq:2}. \begin{align}
f(\text{loss})  &= \| \mathbf{H} \mathbf{x} - \mathbf{y} \|^2= (\mathbf{H} \mathbf{x} - \mathbf{y})^{H} (\mathbf{H} \mathbf{x} - \mathbf{y})\\
&= \mathbf{x}^{H} \mathbf{H}^{H} \mathbf{H} \mathbf{x} - \mathbf{y}^{H} \mathbf{H} \mathbf{x} - \mathbf{x}^{H} \mathbf{H}^{H} \mathbf{y} + \mathbf{y}^{H} \mathbf{y}
\label{eq:2}.
\end{align}

The optimization function includes 2 steps : 1. a gradient descent (GD) method; 2, a dynamic partially shrinkage thresholding (DPST) method. We utilized Wirtinger calculus \cite{amin2011wirtinger}, which allows differentiation of complex-valued functions with respect to complex variables. The special partial derivatives, called Wirtinger derivatives, which simplify calculations by treating complex variables and their conjugates as independent. The Wirtinger derivative of a complex function $\partial f(z)$ of a complex variable $z = x + iy \in \mathbb {C} , x,y \in \mathbb {R} $, with respect to $z$ and $ z^{H} $ (hermitian martrix of $z$), are defined as equations ~\ref{eq:1}.
\begin{align}
\frac{\partial f}{\partial z} \triangleq \frac{1}{2} \left( \frac{\partial f}{\partial x} - j \frac{\partial f}{\partial y} \right) & \quad \frac{\partial f}{\partial z^{H} } \triangleq \frac{1}{2} \left( \frac{\partial f}{\partial x} + j \frac{\partial f}{\partial y} \right)
\label{eq:1}
\end{align}
We use the operator of wirtinger derivative with respect to $z^{H}$.
\begin{equation}
\nabla f(x)= \frac{\partial f}{\partial \mathbf{x}^H} = \mathbf{H}^{H} \mathbf{H} \mathbf{x} - \mathbf{H}^{H} \mathbf{y} = \mathbf{H}^{H} (\mathbf{H} \mathbf{x} - \mathbf{y})
\label{eq:4}
\end{equation}

The dynamic partially shrinkage thresholding (DPST) solution includes the following steps:
 \begin{equation}
\mathbf{x}_{t+1} = \mathbf{x}_t - \gamma_{t}\mathbf{H}^{H} (\mathbf{H} \mathbf{x_{t}} - \mathbf{y})
\label{eq:6}
\end{equation}
\begin{equation}
x_{t+1} = 
\begin{cases} 
      |\theta| \tanh(x_{t+1}) & \text{if } t \geq p\times T, \\
      x_{t+1} & \text{otherwise}.
   \end{cases}
\end{equation}

Let \(t =\{1, \ldots, T\} \), and \( \tanh(\cdot) \) serve as a shrinkage function. A threshold parameter \( p \) (\( 0 < p < 1 \)) is used to control the layer at which the shrinkage function is applied. Specifically, if the current iteration \( t \) exceeds \( p \times T \), the shrinkage function is activated. The parameters \( \{\gamma_{t}\}_{t=1}^{T} \) and \( \{\theta_{t}\}_{t=1}^{T} \) are learnable and can be optimized via backpropagation. These parameters govern the behavior of each layer in the network and adapt the model dynamically during training, guided by the threshold-based shrinkage mechanism.

\section{Experiment and Evaluation}\label{experiment}

In our experimental setup, we consider a MIMO system with a transmitter equipped with \( N_t = 4 \) antennas and a receiver equipped with \( N_r = 8 \) antennas, forming a MIMO configuration. Quadrature Amplitude Modulation (QAM) is employed as the modulation scheme, with a modulation order of 2. The system performance is evaluated over a range of signal-to-noise ratio (SNR) values: \( \{0, 5, 10, 15, 20, 25\} \) dB; T values of  \( \{10, 20, 30, 50, 100\} \). Each simulation is performed over 10{,}000 iterations to ensure statistical significance, and data is processed in batches with a batch size of 24.

\begin{figure}[t] 
  \centering
  \begin{subfigure}[b]{0.48\linewidth}
    \centering
    \includegraphics[height=1.5in, width=\linewidth]{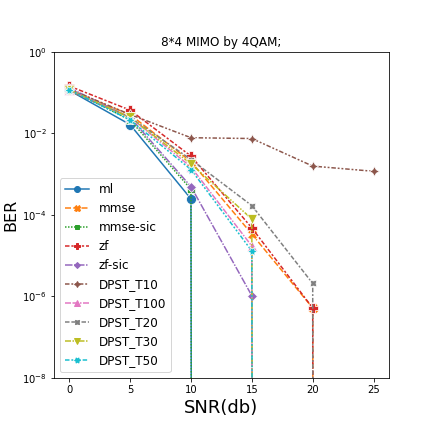}
    \caption{BER performance}
    \label{fig:ber1}
  \end{subfigure}
  \hfill
  \begin{subfigure}[b]{0.46\linewidth}
    \centering
    \includegraphics[height=1.3in, width=\linewidth]{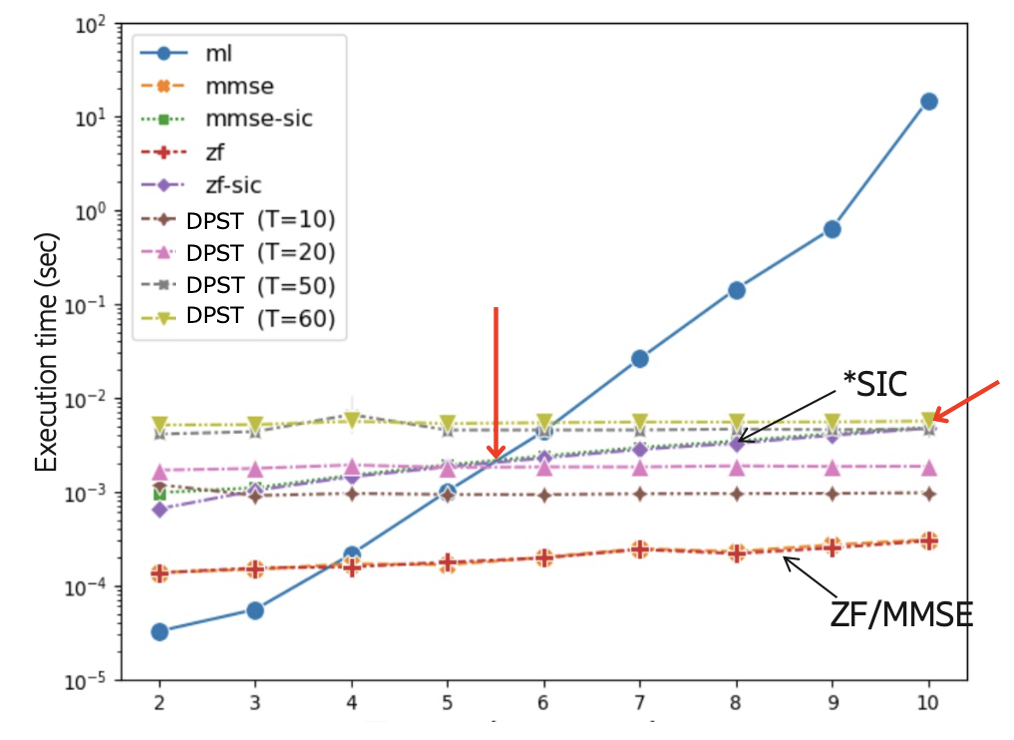}
    \caption{Execution time}
    \label{fig:time2}
  \end{subfigure}
  \caption{Comparison of detection performance under various settings}
  \label{fig:ber_combined}
\end{figure}

As shown in Figure~\ref{fig:ber1} and Figure~\ref{fig:time2}, the Maximum Likelihood (ML) detector achieves the lowest BER across the tested SNR range, but at a high computational cost. It increases exponentially with MIMO system size. Other linear detectors such as MMSE and ZF show moderate performance, while their SIC-enhanced versions (MMSE-SIC and ZF-SIC) exhibit significant BER improvements at high SNR levels. The proposed DPST-based methods demonstrate competitive BER performance, especially as the training length increases. DPST\_T100, for instance, nearly matches the performance of ML in the high-SNR region, confirming the effectiveness of embedding-based learning and temporal optimization. Even with shorter training durations (e.g., T30, T50), the BER is significantly reduced compared to standard linear detectors, indicating the scalability and adaptability of the DPST algorithm.


The DPST-based detectors strike a favorable balance between detection performance and computational complexity. While longer training durations slightly increase computational time, the overall execution remains significantly lower than ML and even competitive with SIC-based methods. Notably, DPST\_T20 and DPST\_T30 achieve strong BER results with execution times comparable to MMSE and ZF, suggesting practical applicability in low-latency environments.


\section{Conclusion}
In this paper, we proposed a novel MIMO signal detection algorithm based on a deep unfolding neural network architecture operating in the complex domain. Leveraging Wirtinger calculus for complex-valued gradient computation, the proposed DPST method enables an efficient and interpretable optimization process for MIMO detection. The numerical experiments demonstrated that DPST achieves BER performance comparable to maximum likelihood (ML) detection while significantly reducing computational complexity.



\bibliography{IEEEabrv,aaai22}

\bibliographystyle{IEEEtran}

\vspace{12pt}

\end{document}